# Deep Prior

**Alexandre Lacoste**
allac@elementai.com

**Thomas Boquet**
thomas@elementai.com

**Negar Rostamzadeh**
negar@elementai.com

**Boris Oreshkin**
boris@elementai.com

**Wonchang Chung**
wonchang@elementai.com

**David Krueger**
david.krueger@elementai.com

## Abstract

The recent literature on deep learning offers new tools to learn a rich probability distribution over high dimensional data such as images or sounds. In this work we investigate the possibility of learning the prior distribution over neural network parameters using such tools. Our resulting variational Bayes algorithm generalizes well to new tasks, even when very few training examples are provided. Furthermore, this learned prior allows the model to extrapolate correctly far from a given task's training data on a meta-dataset of periodic signals.

## 1 Learning a Rich Prior

Bayesian Neural Networks [1, 2, 3, 4] are now scalable and can be used to estimate prediction uncertainty and model uncertainty [5]. While many efforts focus on better approximation of the posterior, we believe that the quality of the uncertainty highly depends on the choice of the prior. Hence, we consider learning a prior from previous tasks by learning a probability distribution $p(w|\alpha)$ over the weights $w$ of a network, parameterized by $\alpha$, and leveraging this learned prior to reduce sample complexity on new tasks.

More formally we consider a hierarchical Bayes approach across $N$ tasks, with hyper-prior $p(\alpha)$. Each task has its own parameters $w_j$, with $\mathcal{W} = \{w_j\}_{j=1}^N$. Using all datasets $\mathcal{D} = \{S_j\}_{j=1}^N$, we have the following posterior:[1]

$$p(\mathcal{W}, \alpha | \mathcal{D}) = p(\alpha | \mathcal{D}) \prod_j p(w_j | \alpha, S_j)$$
$$\propto p(\mathcal{D}|\mathcal{W})p(\mathcal{W}|\alpha)p(\alpha)$$
$$\propto \prod_j \prod_i p(y_{ij}|x_{ij}, w_j)p(w_j|\alpha)p(\alpha),$$

For simplicity, in this work, we consider a point estimation of $p(\alpha|\mathcal{D})$. This can be justify by considering scenarios where we have a lot of samples to learn $\alpha$ across many tasks while the uncertainty we truly care about is the uncertainty over $w_j$ for new tasks.

To go beyond normal distributions for expressing $p(w_j|\alpha, S_j)$, we get inspiration from the generator of generative adversarial networks [6]. We use an auxiliary variable $z \sim \mathcal{N}(0, I)$ and a deterministic function projecting the noise $z$ to the space of $w$ i.e. $w = h_\alpha(z)$. Marginalizing $z$, we have: $p(w|\alpha) = \int_z p(z)p(w|z,\alpha)dz = \int_z p(z)\delta_{h_\alpha(z)-w}dz$, where $\delta$ is the Dirac delta function. Unfortunately, directly marginalizing $z$ is untractable for general $h_\alpha$. To overcome this issue, we add $z$ to the joint

---

[1]$p(x_{ij})$ cancelled with itself from the denominator since it does not depend on $w_j$ nor $\alpha$. This would have been different for a generative approach.

inference and we marginalize at inference time. Considering the point estimation of $\alpha$, we now have:

$$\prod_{j=1}^{N} p(w_j|z_j,\alpha,S_j)p(z_j|\alpha,S_j) \propto \prod_{j=1}^{N} p(w_j|z_j,\alpha)p(z_j)\prod_{i=1}^{n_j} p(y_{ij}|x_{ij},w_j),$$

where $p(y_{ij}|x_{ij},w_j)$ is simply the conventional likelihood function of a neural network with weight matrices generated from the function $h_\alpha$ i.e.: $w_j = h_\alpha(z_j)$. Also, we use:

$$p(z_j) = \mathcal{N}(0,I)$$
$$p(z_j,w_j|\alpha) = p(z_j)\delta_{h_\alpha(z_j)-w_j}$$
$$p(z_j,w_j|\alpha,S_j) = p(z_j|\alpha,S_j)\delta_{h_\alpha(z_j)-w_j}$$

The task now consists in jointly learning a function $h_\alpha$ common to all tasks and a posterior distribution $p(z_j|\alpha,S_j)$ for each task. At inference time, predictions are performed by marginalizing $z$ i.e.: $p(y|x,\mathcal{D}) = \mathbb{E}_{z \sim p(z_j|\alpha,S_j)} p(y|x,h_\alpha(z))$

## 1.1 Hierarchical Variational Bayes Neural Network

Given a family of distributions $q(\{w_j,z_j\}_{j=1}^N) = \prod_j q_{\theta_j}(z_j|S_j)\delta_{h_\alpha(z_j)-w_j}$, parameterized by $\{\theta_j\}_{j=1}^N$ and $\alpha$, the Evidence Lower Bound (ELBO) is:

$$\ln p(\mathcal{D}) \geq \mathbb{E}_{q(\{w_j,z_j\}_{j=1}^N)} \sum_{j=1}^{N}\sum_{i=1}^{n_j} \ln p(y_{ij}|x_{ij},w_j) - \mathrm{KL}\left(q \parallel p\right),$$

$$= \sum_{j=1}^{N} \mathbb{E}_{q_{\theta_j}(z_j|S_j)} \sum_{i=1}^{n_j} \ln p(y_{ij}|x_{ij},h_\alpha(z_j)) - \sum_j \mathbb{E}_{q(w_j,z_j)} \ln \frac{q_{\theta_j}(z_j|S_j)}{p(z_j)}\frac{\delta_{h_\alpha(z_j)-w_j}}{\delta_{h_\alpha(z_j)-w_j}}$$

$$= \sum_{j=1}^{N} \mathbb{E}_{q_{\theta_j}(z_j|S_j)} \sum_{i=1}^{n_j} \ln p(y_{ij}|x_{ij},h_\alpha(z_j)) - \sum_j \mathrm{KL}\left(q_{\theta_j}(z_j|S_j) \parallel p(z_j)\right)$$

Note that after simplification[2], the ELBO no longer depends explicitly on $w$. This is due to the fact that both the posterior and the prior are using the same function to project $z$ into $w$ space[3]. A similar simplification also happened in the likelihood, leaving the whole loss function independent of $w$. This simplification has a major positive impact on the scalability of our approach. Since we no longer need to explicitly calculate the *KL* on the space of $w$, we can simplify the likelihood function to the following $p(y_{ij}|x_{ij},z_j,\alpha)$, which can be a deep network, parameterized by $\alpha$ taking both $x_{ij}$ and $z_j$ as inputs. This contrasts with the previous formulation where $h_\alpha(z)$ produces all the weights of a network, yielding a really high dimensional representation and slow training.

## 2 Related Work

To our knowledge, the only existing work performing hierarchical Bayesian inference with neural networks for multi-task and few shot learning is the Neural Statistician [7]. Their algorithm shares important similarities with ours. They have a main network conditioned on a probabilistic encoding of the task, learned through variational Bayes. However, the main network is a generative model instead of a discriminative model i.e. they use a variational auto-encoder [8]. Also, instead of using a free form embedding for each task, they use an encoding network which reads every samples in the training set to generate the task encoding. Those two differences make our algorithm simpler to implement and more scalable to bigger dataset. Finally, their algorithm was not developed under the perspective of weight uncertainty and there is no exploration of model uncertainty in their experiments.

Some recent papers on meta-learning are also targeting transfer learning from multiple tasks. Model-Agnostic Meta-Learning [9] finds a shared parameter $\theta$ such that for a given task, one gradient step on

---
[2]We can justify the cancellation of the Dirac delta functions by instead considering a Gaussian with finite variance, $\epsilon$. For all $\epsilon > 0$, the cancellation is valid, so letting $\epsilon \to 0$, we recover the result.

[3]Since the posterior need to stay within the support of the prior, there is no need to do otherwise

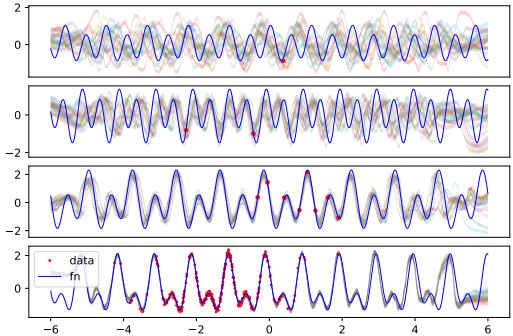
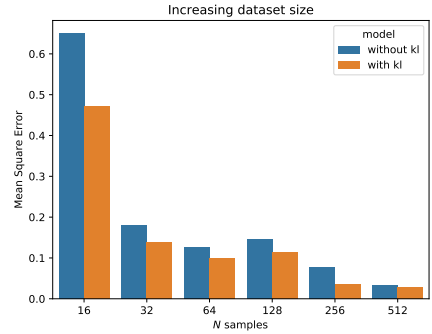

Figure 1: Preview of a few tasks (blue line) with increasing amount of training samples (red dots). Samples from the posterior distribution are shown in semi-transparent colors. The width of each samples is two standard deviations (provided by the predicted heteroskedastic noise).

Figure 2: Mean Square Error on increasing dataset size. The baseline corresponds to the same model without the KL regularizer. Each value is averaged over 100 tasks and 10 different restart.

$\theta$ using the training set will yield a model with good predictions on the test set. Then, a meta-gradient update is performed from the test error through the one gradient step in the training set, to update $\theta$. This yields a simple and scalable procedure which learns to generalize. However this approach does not enable model uncertainty. Finally, [10] also considers a meta-learning approach where an encoding network reads the training set and generate the parameters of a model, which is trained to perform well on the testings set.

## 3 Experimental Results

### 3.1 Regression on 1d Harmonic signals

We demonstrate the ability of our model to learn a good prior on a dataset of periodic signals. The model successfully generalizes the periodic structure to unseen signals, while maintaining appropriate uncertainty about *which* periodic signal the data is sampled from. Specifically, each dataset consists of $(x, y)$ pairs (noisily) sampled from a sum of two sine waves with different phase and amplitude (but the same frequency):

$$f(x) = a_1 \sin(\omega \cdot x + b_1) + a_2 \sin(2 \cdot \omega \cdot x + b_2); \quad y \sim \mathcal{N}(f(x), \sigma_y^2).$$

We construct a meta-training set of 5000 tasks, sampling $\omega \sim \mathcal{U}(5,7)$, $(b_1, b_2) \sim \mathcal{U}(0, 2\pi)^2$ and $(a_1, a_2) \sim \mathcal{N}(0,1)^2$ independently for each task. Then, $x$ values are sampled according to $\mathcal{N}(\mu_x, 1)$ where $\mu_x \sim \mathcal{U}(-4, 4)$ and the number of training samples ranges from 4 to 50. Evaluation is performed on tasks never seen during training.

### 3.2 Model

For a simple implementation of $p(\boldsymbol{z}_j|S_j)$, one can use $\mathcal{N}(\boldsymbol{\mu}_j, \boldsymbol{\sigma}_j^2)$, where $\boldsymbol{\mu}_j$ and $\boldsymbol{\sigma}_j$ are $d$ dimensional vectors for each task $j$, learned through the reparameterization trick [8]. But, we found that inverse autoregressive flow (IAF) [11] converges at a faster rate and yields better final results.

Once $z$ is obtained, we simply concatenate with $x$ and use 12 densely connected layers of 128 neurons with residual connections between every other layer. The final layer linearly projects to 2 outputs $\mu_y$ and $s$, where $s$ is used to produce a heteroskedastic noise, $\sigma_y = \text{sigmoid}(s) \cdot 0.1 + 0.001$. Finally, we use $p(y|x,z) = \mathcal{N}(\mu_y, \sigma_y^2)$ to express the likelihood of the training set. To help gradient flow, we use ReLU activation functions and Layer Normalization[4] [12].

---

[4]Layer norm only marginally helped.

### 3.3 Discussion

The results in Figure 1 exhibit the expected behavior. In the first plot, we can see the posterior latching on the single training point and, most importantly, we observe a wide diversity of different functions plausible under what we consider a good learned prior. In the second plot, with 2 training points, the posterior is already having a good idea of the frequency. Then with 8 points, the task is mostly solved and we are able to extrapolate far away from the training points. Finally with 256 points, the posterior become highly confident on the observed points, and leverages the learned "periodic function" prior to extrapolate confidently in regions without any observations.

In Figure 2, we compare the mean squared error against a version of the model with no *KL*, which would correspond to a more traditional approach to multi-task learning. The comparison is performed on an increasing size of the training set and we can observe a systematic and significant gain over the baseline. We also compared using the log likelihood metric and the Without-KL version is off the chart, performing 10 times worse on average.

We also observed some local minimum issues during training on new tasks[5]. We believe it is caused by the posterior distribution being highly multi-modal in the small sample size regime[6]. In principle, IAF should be able to handle multi-modal distributions. However, extended experiments showed it was not adapting beyond a single distorted mode.

---

[5] Which was resolved with a few restart selected on a validation set.

[6] With more data, the main mode highly shadows the other modes